\documentclass[sigconf]{acmart}

\AtBeginDocument{%
  }

\copyrightyear{2024}
\acmYear{2024}
\setcopyright{acmlicensed}\acmConference[ICMI '24]{INTERNATIONAL CONFERENCE ON MULTIMODAL INTERACTION}{November 4--8, 2024}{San Jose, Costa Rica}
\acmBooktitle{INTERNATIONAL CONFERENCE ON MULTIMODAL INTERACTION (ICMI '24), November 4--8, 2024, San Jose, Costa Rica}
\acmDOI{10.1145/3678957.3685732}
\acmISBN{979-8-4007-0462-8/24/11}

\begin{document}

\title{Mitigation of gender bias in automatic facial non-verbal behaviors generation}

\author{Alice Delbosc}
\affiliation{%
  \institution{Davi, The Humanizers}
  \city{Puteaux}
  \country{France}
}
\additionalaffiliation{Aix-Marseille Univ, CNRS, LIS, Marseille, France}
\additionalaffiliation{Université Paris-Saclay, CNRS, LISN, Orsay, France}
\email{alice.delbosc@lis-lab.fr}

\author{Magalie Ochs}
\affiliation{%
  \institution{Aix-Marseille Univ, CNRS, LIS}
  \city{Marseille}
  \country{France}
}
\email{magalie.ochs@lis-lab.fr}

\author{Nicolas Sabouret}
\affiliation{%
  \institution{Université Paris-Saclay, CNRS, LISN}
  \city{Orsay}
  \country{France}
}
\email{nicolas.sabouret@universite-paris-saclay.fr}

\author{Brian Ravenet}
\affiliation{%
  \institution{Université Paris-Saclay, CNRS, LISN}
  \city{Orsay}
  \country{France}
}
\email{brian.ravenet@limsi.fr}

\author{Stéphane Ayache}
\affiliation{%
  \institution{Aix-Marseille Univ, CNRS, LIS}
  \city{Marseille}
  \country{France}
}
\email{stephane.ayache@lis-lab.fr}

\renewcommand{\shortauthors}{Delbosc, et al.}

\begin{abstract}
Research on non-verbal behavior generation for social interactive agents focuses mainly on the believability and synchronization of non-verbal cues with speech. However, existing models, predominantly based on deep learning architectures, often perpetuate biases inherent in the training data. This raises ethical concerns, depending on the intended application of these agents. This paper addresses these issues by first examining the influence of gender on facial non-verbal behaviors. We concentrate on gaze, head movements, and facial expressions. We introduce a classifier capable of discerning the gender of a speaker from their non-verbal cues. This classifier achieves high accuracy on both real behavior data, extracted using state-of-the-art tools, and synthetic data, generated from a model developed in previous work.
Building upon this work, we present a new model, \textit{FairGenderGen}, which integrates a gender discriminator and a gradient reversal layer into our previous behavior generation model. This new model generates facial non-verbal behaviors from speech features, mitigating gender sensitivity in the generated behaviors. Our experiments demonstrate that the classifier, developed in the initial phase, is no longer effective in distinguishing the gender of the speaker from the generated non-verbal behaviors.
\end{abstract}



\begin{teaserfigure}
  \includegraphics[width=\textwidth]{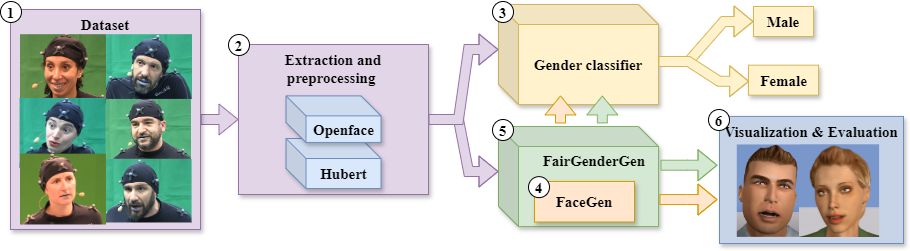}
    \caption{Overview of the fair gender generative model: {\normalfont\it (1) From a video corpus, (2) extraction of verbal and non-verbal features. (3) Classifier to verify the feasibility of gender identification based on non-verbal features extracted from the corpus. (4) Evaluation of the gender identification of a facial generation model based on an adversarial approach (\textit{FaceGen}). (5) Introduction of a model to mitigate the gender bias in facial generation (\textit{FairGenderGen}). (6) Comparison of the generated behavior of the two models (\textit{FaceGen} and \textit{FairGenderGen}) through objective and subjective studies, employing various SIAs.}}
  \Description{Overview of the fair gender generative model, described in detail in the article.}
  \label{fig:teaser}
\end{teaserfigure}


\maketitle

\vspace*{-1mm}
\section{Introduction} \label{intro}
Socially Interactive Agents (SIAs) are virtual agents that simulate key properties of face-to-face human conversation, such as verbal and non-verbal behaviors. A number of studies have been carried out to simulate role-playing with SIAs to train one's own skills \cite{bruijnes2019special}, for example for training doctors to break bad news \cite{ochs2019training}, job interviews \cite{anderson2013tardis}, negotiation \cite{gratch2016benefits}, or conflict management \cite{lee_who_2021}. A crucial aspect for the widespread acceptance and use of these applications lies in the believability of the non-verbal behaviors exhibited by the SIAs.
The SIAs' non-verbal behavior is particularly important, since several studies underline the positive impact of non-verbal behaviors on knowledge transmission and information retention \cite{davis2018impact}. In addition, studies indicate that appropriate head movements enhance the overall perception of SIAs, while inappropriate facial expressions can increase their sense of “uncanniness” \cite{tinwell2011facial}. Early approaches explored for the automatic generation of SIA's behaviors were based on sets of rules \cite{cassell1994animated, kopp2002model}. The rules describe the mapping of words or speech to a facial expression or gesture. These approaches present advantages in terms of communication and control but lack naturalness and variability in behavior generation \cite{nyatsanga2023comprehensive}. Nowadays, most of the research works on behavior generation are based on data-driven approaches \cite{habibie2021learning}. These approaches do not depend on experts in animation and linguistics. They learn the relationships between speech and movements, or facial expressions, directly from data. Among data-driven approaches, deep neural networks have demonstrated their superiority in this task. The commonly employed methodological approach is to extract verbal and non-verbal features from recorded real-world human interactions, and to train a generative model using these real-world datasets \cite{kucherenko2021moving, habibie2021learning, haque2023facexhubert, delbosc2023towards}. Two key aspects are often evaluated to determine the performance of these models: the human-likeness and the appropriateness of the non-verbal behaviors with speech \cite{kucherenko2023evaluating}. However, the presence of possible bias in such models is rarely considered a criterion for evaluating the quality of the model.

Indeed, real-world datasets are often biased \cite{choi2020fair}. The most frequently identified biases come from key demographic factors like gender. We know, for instance, that men and women differ in their non-verbal behaviors \cite{lafrance2016gender}. As generative models learn from our data, most contain biases by simply reproducing those that have been passed to them \cite{frankel2020fair}. This may raise ethical concerns, depending on the intended use of these agents. While it might be wanted to reproduce societal norms and behaviors in SIAs, \textit{e.g.} for better cultural understanding and acceptability, reproducing gender biases can perpetuate harmful stereotypes and inequalities, contributing to the normalization of discriminatory attitudes and behaviors in society. Indeed, a recent study shows that humans inherit the biases of the artificial intelligence they use \cite{vicente2023humans}. Moreover, biased SIAs may make users who don't conform to traditional gender norms or identities feel marginalized or different. Allowing SIAs to perpetuate biases raises ethical questions about the responsibilities of technology creators to promote fairness, equity, and inclusion.

This work addresses this issue of bias and fairness in models of facial non-verbal behavior generation, focusing on gender bias. In the field of generative models, fairness is generally defined as equal generative quality or equal representation \cite{teo2023fair}, for instance of men and women in generated images. In our context, we define fairness as the absence of distinction in the generated non-verbal behaviors, whatever the gender of the speaker. We aim to avoid the perpetuation of gender stereotypes and biases in non-verbal behavior, by adopting an approach in which the generated non-verbal behaviors are not differentiated according to gender. In this article, as a first step, we focus on the gaze, the head movements and the facial expressions. The research questions addressed are: 
\begin{itemize}
    \item Do generative models reproduce potential differences in non-verbal behavior between the genders?
    \item Can we modify the model to mitigate the gender differences in non-verbal behavior generation without compromising the perceived naturalness and appropriateness of these behaviors with speech?
\end{itemize}

The paper is organized as follows: we first provide an overview of existing works on fairness in generative models in Section \ref{state}. The corpus and feature extraction are presented in Section \ref{dataset}. The baseline generative model is introduced Section \ref{FaceGen}. The gender classifier and the results of the classification are described in Section \ref{classifier}. Section \ref{FairGenderGen} is devoted to the architecture of the \textit{FairGenderGen} model to generate non-verbal behaviors with mitigation of gender bias. Section \ref{eval} is dedicated to the evaluation of the models. Finally, we conclude the paper and introduce perspectives in Section \ref{conclusion}. The workflow is illustrated Figure \ref{fig:teaser}.

\vspace*{-1mm}
\section{Related work} \label{state}
Research on bias and fairness has a long history in philosophy, psychology, and in recent years in machine learning \cite{mehrabi2021survey}. While machine learning ethics often focuses on classification problems, such as gender-neutral hiring \cite{teo2023fair}, recent attention has turned towards the ethical implications of generative models \cite{frankel2020fair, choi2020fair, liu2015deep}. 

For these models to be practically viable, they must meet ethical standards and be free from biases that may perpetuate human prejudices. This work contributes to the ethical development of SIAs by addressing gender bias in this domain. To our knowledge, no other research on automatic non-verbal behavior generation has addressed such an ethical dimension. 

To effectively rectify these biases and achieve fairness, it is imperative to first establish clear definitions of what constitutes fairness and identify existing discrimination.

\vspace*{-1mm}
\subsection{Definition of fairness} \label{state:fairness}
The definition of \textit{fairness} varies according to the context in which it is applied. Some definitions of fairness focus on \textit{equal representations} of certain sensible attributes, for example, a generative model that has equal probabilities of producing male or female examples \cite{teo2023fair}. 

In our study, we focus on the generation of non-verbal behaviors from speech. While biases in non-verbal behaviors can be addressed by balancing sensitive attributes in datasets, the concept of fairness related to equal gender representation is not our focus. We wish to concentrate on the intrinsic differences in non-verbal patterns exhibited by individuals of different gender identities.

In the context of generative models, some definitions emphasize \textit{performance fairness} \cite{maluleke2022studying}. These approaches seek consistency in generation quality, whatever the sensitive attribute considered, such as gender. Although this approach may apply to our particular situation, generating behaviors with the same performance for men and women in no way ensures that these behaviors do not depend on the gender of the speaker. It is therefore not suitable for working on the generation of non-stereotyped behaviors.

One of the definitions explored in the survey by \citet{mehrabi2021survey} is: “\textit{an algorithm is fair as long as any protected attributes are not explicitly used in the decision-making process}”. We adapt this definition to the generation of non-verbal behaviors and define fairness as “the absence of distinction in the generated non-verbal behaviors, whatever the gender of the speaker”. We aim to avoid the perpetuation of gender stereotypes and biases in the non-verbal behavior and to generate non-verbal behaviors that are not differentiated according to gender.

\vspace*{-1mm}
\subsection{Approaches to mitigate bias} \label{state:approaches}
Biases in model-generated data come mainly from two sources: the dataset and the models themselves. Dataset bias is the main cause of unfairness in generative models. One solution is to work with unbiased data. Practical limitations such as time, resources, and the complex nature of non-verbal behavior, render this approach difficult. We cannot simply balance datasets on the basis of the distribution of sensitive attributes, since bias comes from the fact that individuals have different non-verbal behaviors depending on their gender identity \cite{lafrance2016gender}. 

Models can perpetuate and even amplify biases in the data \cite{choi2020fair}. Generative Adversarial Networks (GANs), for instance, are trained in an unsupervised way to capture the underlying distribution of the dataset, then generate new data from the same distribution \cite{teo2023fair}. 

To address these issues, researchers have explored various techniques, including pre-processing, in-processing, and post-processing \cite{bellamy2018ai}. While pre-processing and post-processing methods directly manipulate data, in-processing approaches modify the model during training. Despite advancements in bias mitigation, there is a paucity of research specifically addressing bias in generated non-verbal behaviors. We investigate this aspect from the perspective of generation models in general.

Pre-processing attempts to transform data to remove distribution bias, and post-processing involves modifying the generated data after the model has been trained. \citet{xu2018fairgan} work with adversarial networks, trying to generate new data free of the discriminant attribute. They generate new datasets similar to real data that are debiased and preserve good data utility. We felt that it was more effective to operate at the learning stage, building a model that learns from our “biased” data “non-biased” non-verbal behaviors whatever the speaker's gender is.

Several methods have been proposed to mitigate biases in generative models with an in-processing approach. In the context of image generation, \citet{choi2020fair, teo2023fair} use a complementary unbiased dataset as a supervisory signal to detect bias in the baseline data and bring the distribution of the baseline data closer to the reference data. In these works, it is assumed that an unbiased dataset can be accessed or constructed. \citet{zhang2018mitigating} employ adversarial learning by presenting a model in which they try to maximize the accuracy of a predictor and at the same time minimize the ability of an adversary to predict the sensitive variable. They use adversarial learning to mitigate sensitive attributes, a method noted by \citet{frankel2020fair} as costly to train. \citet{frankel2020fair} develop a method that uses a small neural network ahead of the existing generator to perturb the latent variables. While this approach effectively addresses fairness, it increases both training and inference times due to the additional network layer.

Similarly, some studies seek to learn latent representations that remain invariant with respect to a given variable. One example is the \textit{Variational Fair Autoencoder} \cite{louizos2015variational}, which extends the semi-supervised variational autoencoder to acquire representations explicitly invariant to known dataset attributes. By employing a Maximum Mean Discrepancy regularizer, they promote invariant latent variable distributions. This approach necessitates the use of a specialized variational encoder architecture. 

The field of domain adaptation, for example the work of \citet{ganin2015unsupervised}, closely relates to this approach by seeking to minimize the discrepancy between feature distributions of two domains. Their findings demonstrate that adaptation can be integrated into nearly any feed-forward model by adding a small set of standard layers along with a novel gradient reversal layer. Unlike previous methods, this technique enables iterative training, reducing computational costs, and the gradient reversal layer is only active during training, not affecting inference. Furthermore, this approach permanently modifies the latent representation, eliminating the need for an additional neural network before the generator. We propose adapting the approach of \citet{ganin2015unsupervised} to mitigate gender bias in non-verbal behavior generation. 

\vspace*{-1mm}
\section{Facial behaviors corpus} \label{dataset}
Focusing on the automatic generation of facial expressions, head movements and gaze, a corpus that emphasizes facial recordings with a balanced representation of male and female speakers is required. For this purpose, we use the \textit{Trueness} corpus \cite{ochs2023forum}.

\vspace*{-1mm}
\subsection{Presentation and splitting}
\textit{Trueness} is a corpus of scenes of ordinary discrimination, of sexism and of racism \cite{ochs2023forum}. It also includes interactions between authors of discriminatory behavior and witnesses, attempting to sensitize them by acting out various socio-affective behaviors such as aggression, conciliation or denial. These scenes originate from a French forum theater focused on discrimination, with professional actors trained in this domain. Each scene is divided into two videos, representing the perspectives of the first and second persons in the interaction. An essential quality aspect of the facial non-verbal behaviors is the camera's field of view, carefully maintained to capture only the face and torso.

The dataset is divided into two parts, each recorded separately with different actors. The first part comprises a training set, \textit{SetGen}, and a test set, \textit{TestSet}, used for training and evaluating generative models. The second part, \textit{SetClassif}, is dedicated to train the gender classifier. To ensure dataset diversity and prevent data overlap, individuals are exclusively included in either \textit{SetGen}, \textit{SetClassif}, or \textit{TestSet}. Specifically, \textit{SetClassif} contains approximately 4 hours and 30 minutes of recordings from two male and two female speakers. \textit{SetGen} includes about 2 hours and 57 minutes of recordings from two male and two female speakers. \textit{TestSet} comprises around 41 minutes of recordings, featuring one male and one female speaker. 

\vspace*{-1mm}
\subsection{Extraction and processing}
We automatically extract behavioral features from the existing videos using \textit{Openface} \cite{baltruvsaitis2016openface} and speech features using the self-supervised speech model \textit{Hubert} \cite{hsu2021hubert}. 

\vspace*{-1mm}
\subsubsection*{\textbf{Behavioral features}}
\textit{Openface} extracts, among others, 28 features characterizing the head, gaze, and facial behaviors of a person on a video at a frequency of 25 fps (frames per second). The eye gaze position is represented in world coordinates, the eye gaze direction in radians, the head rotation in radians, and 17 facial action units in intensity from 1 to 5 (AU01-02, AU04-07, AU09-10, AU12, AU14-15, AU17, AU20, AU23, AU25-26, AU45) based on the Facial Action Coding System \cite{ekman1978facial}. We point out that these features are designed to capture non-verbal facial behaviors, but do not offer precise lip-synchronization.

To ensure that our model learns from clean, plausible data, we filter out images that have been incorrectly processed by \textit{OpenFace}. These include images in which faces are obscured by hands or hair. We then interpolate the transitions between the remaining images. In addition, two further processing steps are applied to head and gaze features. Firstly, the features are smoothed using a median filter with a window size of 7. Secondly, the head and gaze coordinates are centered to ensure that the SIA is facing the user. Finally, as our focus in this project is solely on generating speaking behaviors (and not listening behaviors), we set the behavioral features to zero when the protagonist is not speaking.

These features, noted $F_{b} \in \mathbf{R}^{28}$, are used for the training. $F_{b}$ consists of $F_{head}$, $F_{gaze}$, and $F_{AU}$, representing respectively head movements, gaze orientation, and facial expressions.

\vspace*{-1mm}
\subsubsection*{\textbf{Speech features}}
Drawing on \citet{haque2023facexhubert} work on non-verbal facial behavior generation, we use \textit{Hubert} to extract the speech features. In response to various analyses of different layers of self-supervised speech models \cite{pasad2021layer, pasad2023comparative}, we compare the model's objective performances using different layers of \textit{Hubert} (Section \ref{eval} for more details on the computation of objective performances), and choose to use the twelfth layer to extract the speech features. In \textit{Hubert}, speech features are extracted at a frequency of 50 fps. The speech features extracted from human speech are noted $F_{s} \in \mathbf{R}^{1024}$.

\vspace*{-1mm}
\subsubsection*{\textbf{Sliding window}}
Human behaviors are primarily generated by analyzing short segments with a sliding window approach, spanning from seconds to minutes, based on the socio-emotional phenomena studied \cite{murphy2021capturing}. We segment the videos into 4-second segments with a 0.4-second overlap. Since the speech data has a frame rate of 50 fps, and the behavior data has a frame rate of 25 fps, we use a speech segment length of 200 frames and a behavior segment length of 100 frames, they are aligned during training.

This segmentation process yields 3590 segments for \textit{SetClassif}, comprising 1429 female, 1459 male, and 702 silent segments (where no speech occurs within the 4-second window). \textit{SetGen} consists of 2940 segments, including 1352 female, 1002 male, and 586 silent segments. \textit{TestSet} comprises 676 segments, divided into 267 female, 344 male, and 65 silent segments. No segment appears in more than one set.

The extracted and processed data, forming \textit{SetClassif}, \textit{SetGen}, and \textit{TestSet}, are part of the \textit{ground truth} data. These datasets underpin all the processes outlined in Figure \ref{fig:teaser}. We develop a behavior generation model using standard techniques and an existing model, \textit{FaceGen} (detailed in Section \ref{FaceGen}). This model is refined to address gender bias, resulting in \textit{FairGenderGen} (described in Section \ref{FairGenderGen}). \textit{SetGen} is employed for training both \textit{FaceGen} and \textit{FairGenderGen}. The gender classifier (described in Section \ref{classifier}) is trained using \textit{SetClassif}. It is used for inference on \textit{ground truth} data from \textit{TestSet}, as well as data generated by both \textit{FaceGen} and \textit{FairGenderGen} (using speech features from \textit{TestSet}).

\vspace*{-1mm}
\section{The \textit{Face-Gen} model} \label{FaceGen}
We aim to generate non-verbal facial behaviors for SIAs while they speak. We can formulate the goal as follows: given a set of speech features $F_{s}[0:T]$, taken from a particular speech segment at constant frame intervals of length $T=200$, the goal is to generate the sequence of behaviors $Y_{b}[0:\frac{T}{2}]$ that a SIA is expected to perform during its speech. The distribution of $Y_{b}$ must be as close as possible to the one of $F_{b}$.

Our model is build upon the work of \citet{delbosc2023towards}, who introduced an open-source framework for the automatic generation of non-verbal facial behaviors using action units. We implemented a number of adjustments, including: the audio features extracted with \textit{Hubert}, the reduction of discriminator capacity, the noise formation, and the model hyperparameters. In this section, we present this transformed architecture\footnote{\url{https://github.com/aldelb/FairGenderGen}}. This model will serve as the reference “biased” generative model, which we call \textit{FaceGen}.

\vspace*{-1mm}
\subsection{Architecture}
Like the original model presented in \cite{delbosc2023towards}, \textit{FaceGen} adopts the structure of an adversarial encoder-decoder. It is termed "adversarial" because it comprises two modules, a generator and a discriminator, mirroring the architecture of a GAN \cite{goodfellow2014generative}. The term "encoder-decoder" is employed because the generator operates on the principles of a 1D encoder-decoder. Again, as in the original model, both modules receive speech features $F_{s}[0:T]$, allowing the discriminator to evaluate the believability of the temporal alignment between behavioral and speech features. Preserving this property of the basic model, the discriminator receives (in addition to \textit{ground truth} and generated examples) examples that help it discriminate between speaking and listening phases. These examples associate features of listening behavior with features of speaking, and vice versa. A simplified architecture is shown in the green frame of Figure \ref{archi}. To describe each module, we use the following notations: \textit{Conv} and \textit{DoubleConv}. A \textit{Conv} block is composed of a convolution 1D, dropout, batch normalization 1D, and Relu. A \textit{DoubleConv} block is the concatenation of two \textit{Conv} blocks. 

\vspace*{-1mm}
\subsubsection*{\textbf{The generator}}
The generator generates data by sampling from a noise $z$ and speech features $F_{s}$. The features received by the encoder are not the same as in the original model, so we adapted the architecture, maintaining the main modules. The \textit{encoder} initially learns $F_{s}$ representations using two \textit{Conv} blocks followed by three \textit{DoubleConv} blocks. Each \textit{DoubleConv} is preceded by a maxPool layer. Unlike the original model, this representation is added with noise, not concatenated. The noise is generated by creating two random digits for each channel of $F_{s}$ representation, and using these values to create a noise matching the length of $F_{s}$ representation, with transition digits following one another progressively. Following this, three additional \textit{DoubleConv} blocks, each preceded by a maxPool layer, are applied. The output of the encoder constitutes the latent representation of our data. The \textit{decoder} consists of three decoding modules to generate non-verbal behaviors, each associated with an output type with different value intervals: a decoder for head movements, a decoder for eye movements, and a decoder for AUs. They consist of five \textit{DoubleConv} blocks and an upSampling layer before each. It uses skip-connectivity with the corresponding layers of the encoder. It ends with a convolution 1D and a tanh activation layer.

\vspace*{-1mm}
\subsubsection*{\textbf{The discriminator}}
In parallel to the generator, the discriminator learns separate representations for $F_{s}$ and $F_{b}$. These representations are learned with three $Conv$ blocks for $F_{s}$ and two $Conv$ blocks for $F_{b}$, with a maxPool layer after each block. These representations are then concatenated and processed through one $Conv$ block and two linear layers, followed by a sigmoid activation layer. To enhance computational efficiency without compromising performance, we significantly reduced the network architecture compared to the original model based on evaluation results.

\vspace*{-1mm}
\subsection{Training}
\textit{FaceGen} is optimized with a Wasserstein loss with gradient penalty \cite{gulrajani2017improved}. The generator $G$, with the parameters of the encoder $\theta_e$, and the parameters of the decoders $\theta_d$, is supervised with the following loss function:
\begin{equation*}
\mathcal{L}_G(\theta_{e}, \theta_{d}) = \mathcal{L}_{gaze}(\theta_{e}, \theta_{d}) + \mathcal{L}_{head}(\theta_{e}, \theta_{d}) + \mathcal{L}_{AU}(\theta_{e}, \theta_{d})
\end{equation*}
where $\mathcal{L}_{gaze}$, $\mathcal{L}_{head}$ and $\mathcal{L}_{AU}$ are the root mean square errors (RMSEs) of the gaze orientation, head movement, and AUs features. 

\begin{align*}
\mathcal{L}_{mod}(\theta_{e}, \theta_{d}) &= \sum_{t=0}^{\frac{T}{2}-1}{(F_{mod}[t] - Y_{mod}[t])^2}\\ 
\end{align*}
with $mod\in\{gaze,head,AU\}$. The discriminator $D$, with the parameters $\theta_{a}$, is optimized through the adversarial loss function:
\begin{align*}
L_{adv}(\theta_{e}, \theta_{d}, \theta_{a}) =  & \mathbb{E}_{\tilde{x}\sim\mathbb{P}_{g}}[D(F_{s}, \tilde{x})] -\mathbb{E}_{x\sim\mathbb{P}_{r}}[D(F_{s}, x)] \\ 
& + \phi \underset{\hat{x}\sim\mathbb{P}_{\hat{x}}}{\mathbb{E}}[(||\nabla_{\hat{x}}D(F_{s}, \hat{x})||_2 - 1)^2]
\end{align*}

\noindent with $\mathbb{P}_{r}$ the \textit{ground truth} distribution and $\mathbb{P}_{g}$ the generated distribution defined by $\tilde{x} = G(z, F_{s})$, $z \sim p(z)$. $\mathbb{P}_{\hat{x}}$, used to calculate the gradient norm, samples uniformly between pairs of points sampled from the data distribution $\mathbb{P}_{r}$ and the generator distribution $\mathbb{P}_{g}$, $\hat{x} = (l)F_{b} + (1 - l)Y_{b}$ with $0 \leq l \leq 1$. We use $\phi = 10$. By integrating adversarial loss with direct supervisory loss, our objective is the following: 
\begin{equation*}
\mathcal{L}_y(\theta_{e}, \theta_{d}, \theta_{a}) = \mathcal{L}_G(\theta_{e}, \theta_{d})  + \beta.\mathcal{L}_{adv}(\theta_{e}, \theta_{d}, \theta_{a})
\end{equation*}
we set $\beta=1$. We use Adam optimizer for training, with a learning rate of $10^{-4}$ for the generator and the discriminator. Our batch is size 32. This model was trained for 1200 epochs on a v100 Nvidia GPU, for approximately 14 hours.

\vspace*{-1mm}
\section{Investigating gender bias} \label{classifier}
We assess the presence of gender bias in both \textit{ground truth} and \textit{FaceGen}-generated non-verbal facial behaviors. Following our fairness definition (Section \ref{state:fairness}), a bias is present in non-verbal features if we can identify them as coming from a female or male speaker. For this purpose, we build a \textit{gender classifier}, trained on \textit{SetClassif} (detailed in Section \ref{dataset}). This classifier predicts the speaker's gender based on input non-verbal behavior features, excluding segments of complete silence.

\vspace*{-1mm}
\subsubsection*{\textbf{Architecture and training}}
The gender classifier is a compact neural network composed of two \textit{Conv} blocks (see Section \ref{FaceGen}), each followed by a maxPool operation. Subsequently, there is a linear layer, a ReLU activation function, another linear layer, and finally a log softmax activation layer. The model is trained using cross-entropy loss and the Adam optimizer with a learning rate of $10^{-3}$ for 10 epochs.

\vspace*{-1mm}
\subsubsection*{\textbf{Evaluation and interpretation}}
We train the classifier 10 times to capture variability in the training process, such as random weight initialization and optimization algorithm stochasticity. We train it on a large subset of \textit{SetClassif} (1394 female segments and 1423 male segments) and validate its performances on a smaller subset (35 female segments and 36 male segments). The classifier achieved a mean accuracy of $85.92\%$ with a standard deviation of $4.20\%$. 

Randomly selecting one of the trained classifiers, we classified \textit{ground truth} data from \textit{TestSet}. The resulting accuracy is $90.18\%$ (Table \ref{tab:classif2}) with 4 misclassifications out of 267 for female speakers and 56 misclassifications out of 344 for male speakers. These results indicate that non-verbal behaviors extracted from the dataset exhibit discernible gender patterns, suggesting that gender influences the \textit{ground truth} non-verbal behavior. With this established, we can now explore our initial research question: 'Do generative models reproduce potential differences in non-verbal behavior between the genders?'.

We classified data generated by \textit{FaceGen}. The resulting accuracy is $80.69\%$ (Table \ref{tab:classif2}) with 44 misclassifications out of 267 for female speakers and 74 misclassifications out of 344 for male speakers.The influence of the speaker's gender is evident in both the \textit{ground truth} data and those generated by \textit{FaceGen}. This finding answers our first research question, confirming that gender influence persists in automatically generated behaviors, despite being less pronounced than in \textit{ground truth} data. Therefore, we aim to explore our second research question: "Can we modify the [\textit{FaceGen}] model to mitigate the gender differences in non-verbal behavior generation without compromising the perceived naturalness and appropriateness of these behaviors with speech?". 

\vspace*{-1mm}
\section{The Fair-Gender-Gen model} \label{FairGenderGen}
We introduce a new model called \textit{FairGenderGen}, designed to generate facial non-verbal behaviors from speech, while also aiming to mitigate gender bias by producing behaviors that are independent of the speaker's gender.

\vspace*{-1mm}
\subsection{Architecture}
The model work with speech features $F_{s}[0:T]$ as inputs, and label from the label space \{$female$, $male$, $silence$\}. The approach will nevertheless be generic and can handle any labels. We assume the existence of three distributions: $\mathbb{P}_{f}$, $\mathbb{P}_{m}$ and $\mathbb{P}_{s}$, which will be referred as the \textit{Female}, the \textit{Male} and the \textit{Silence} distributions. All distribution are unknown. We don't deal with the \textit{silence} labels as we aim to maintain the \textit{Silence} distribution unchanged.

Our goal is to achieve a latent representation of our data that is invariant with respect to gender, meaning we aim to make the distributions $\mathbb{P}_{f}$ and $\mathbb{P}_{m}$ as similar as possible. At training time, we have access to labeled examples from both distributions. Measuring the dissimilarity of the distributions is however non trivial as they are consistently changing during the training process.

\begin{figure}[h]
  \centering
  \includegraphics[width=\linewidth]{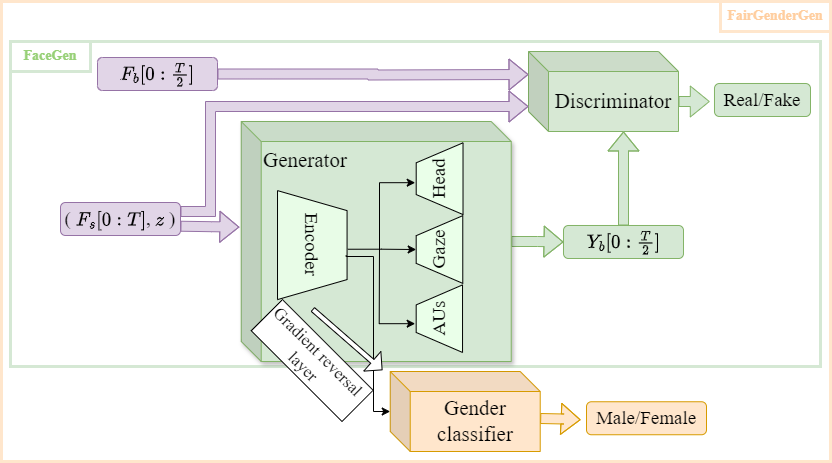}
  \caption{Overall architecture of \textit{FairGenderGen} -- {\normalfont\it The gender classifier (represented in orange) interacting with \textit{FaceGen} (represented in green).}}
  \label{archi}
  \Description{Simplified drawing of the overall architecture of our models, described in detail in the article.}
\end{figure}

Building on prior research presented in Section \ref{state}, we propose to adapt the approach of \citet{ganin2015unsupervised} to mitigate gender bias through domain adaptation with backpropagation. The proposed architecture includes all the modules of the \textit{FaceGen} model (green in Figure \ref{archi}); the generator with encoding and decoding parts, and the discriminator, which together form a standard feed-forward architecture. Unsupervised domain adaptation is achieved by incorporating a gender classifier (orange in Figure \ref{archi}). 

This gender classifier takes as input the latent representation of \textit{FaceGen} data, and classifies them according to the speaker's gender, \textit{male} or \textit{female}. It does not receive the \textit{silent} sequence representations. It is connected to the encoder via a gradient reversal layer. This layer multiplies the gradient by a negative constant during the backpropagation-based training, known as the adaptation factor $\lambda$. Similar to the original paper \cite{ganin2015unsupervised}, we gradually change the adaptation factor from 0 to 1 during the training process.

The gender classifier is a small neural network, consisting of two \textit{Conv} blocks (see Section \ref{FaceGen}), with maxPool after the first block, followed by a linear layer, a ReLU activation function, another linear layer and a log softmax activation layer.  Figure \ref{archi} illustrates the integration of this classifier with the FaceGen model to form the FairGenderGen model.

Gradient reversal ensures that the distributions over the two genders are made as indistinguishable as possible for the gender classifier, thereby resulting in gender-invariant features. This discriminative classifier is only used during training and does not increase inference time. The modified generator operates identically to the original, except with different outputs, modifying the latent variables for a fair generation. 

\vspace*{-1mm}
\subsection{Training}
To avoid starting from scratch and leverage the learning achieved with the \textit{FaceGen} training, we initialize our discriminator and generator with the \textit{FaceGen} weights.

During the learning stage, we optimize the parameters of the encoder $\theta_e$ that maximize the loss of the gender classifier, while simultaneously optimizing the parameters $\theta_c$ that minimize the loss of the gender classifier. The gender classifier uses binary cross-entropy as loss function. Otherwise, the training proceeds in a standard manner, minimizing the overall objective $L_{y}$ with the parameters of the decoders $\theta_d$ and the parameters of the discriminator $\theta_a$. By integrating the loss of the gender classifier $\mathcal{L}_{gender}$ and its parameters $\theta_{c}$, our objective becomes: 
\begin{equation*}
\mathcal{L}_{fair}(\theta_{e}, \theta_{d}, \theta_{a}, \theta_{c})  = \mathcal{L}_{y}(\theta_e, \theta_d, \theta_a) + \alpha.\mathcal{L}_{gender}(\theta_e, \theta_c)
\end{equation*}
with $\alpha$ set to 0.1. We utilize the Adam optimizer for training, with a learning rate of $10^{-4}$ for the generator and the discriminator. Our batch is size 32. This model was trained for 500 epochs on a v100 Nvidia GPU, requiring approximately 6 hours.

\begin{figure*}[h!]
  \centering
  \includegraphics[width=0.8\textwidth]{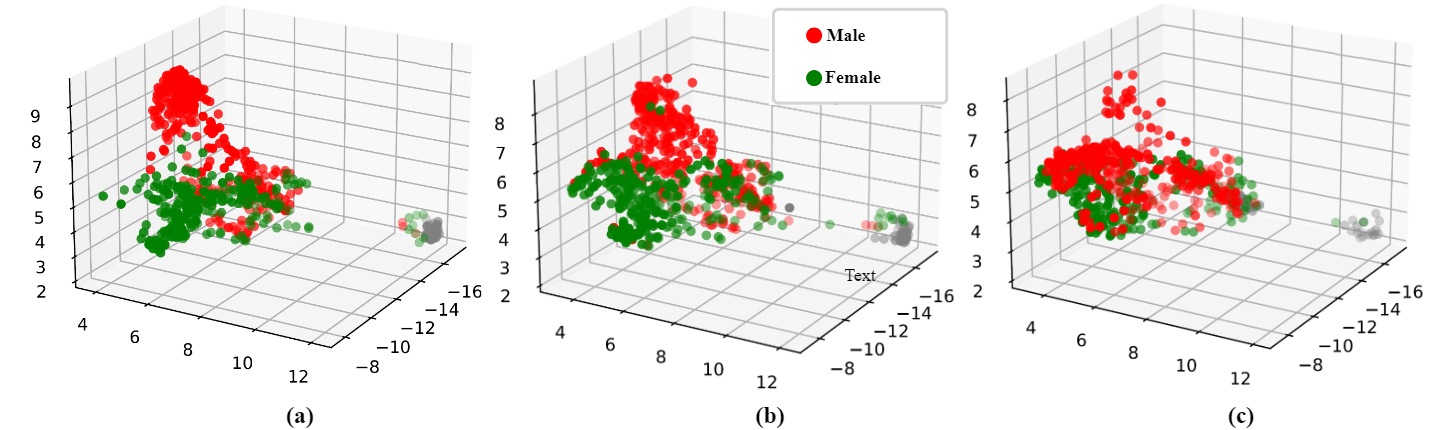}
   \caption{\textit{UMAP} visualization of the non-verbal behaviors - {\normalfont\it{(a) \textit{ground truth} behaviors, (b) \textit{FaceGen}-generated behaviors, and (c) \textit{FairGenderGen}-generated behaviors. Red and green points represent male and female behaviors, respectively.}}}
  \label{umap}
  \Description{UMAP representation of the non-verbal behaviors, described in detail in the article.}
\end{figure*} Figure \ref{umap} displays the generator's outputs on \textit{TestSet} in three dimensions using \textit{UMAP} visualization \cite{mcinnes2018umap}, with the \textit{Male} and \textit{Female} distributions undeniably closer together for the \textit{FairGenderGen} model. To confirm the visualization results showing that the two distributions are closer together, we conduct an objective and subjective evaluation to assess not only the mitigation of bias, but also the consistency of performances (Section \ref{eval}).

\vspace*{-1mm}
\section{Evaluation}\label{eval}
It is equally important to verify that our non-verbal male and female behaviors are now closer, as it is to ensure that the mitigation of bias has not reduced the quality of the generated behaviors.

To address the first point, we use the gender classifier pretrained on the \textit{SetClassif} data (Section \ref{classifier}). This allows us to assess whether the gender differences in non-verbal behaviors have been minimized in the generated data.

For the second point, we \textit{objectively} and \textit{subjectively} evaluate the model's performances. We compare these metrics with those of the \textit{FaceGen} model to ensure that the quality of the generated non-verbal behaviors has been maintained (maintained, improved or slightly degraded).

\vspace*{-1mm}
\subsection{Gender bias}
While the gender classifier (Section \ref{classifier}) was able to discriminate between the non-verbal male and female behaviors generated by the \textit{FaceGen} model with an accuracy of $80.69\%$, its performance significantly dropped to $48.61\%$ when applied to behaviors produced by \textit{FairGenderGen} (Table \ref{tab:classif2}). A closer examination reveals 90 misclassifications out of 267 for female speakers and 224 misclassifications out of 344 for male speakers.

\begin{table}[h]
 \centering
 \setlength\tabcolsep{3pt}
 \caption{Results of the gender classification of \textit{ground truth} behaviors (\textit{Ground truth}), \textit{FaceGen} generated behaviors and \textit{FairGenderGen} generated behaviors -- {\normalfont\it We report results for all features in terms of accuracy (\textit{Acc.}) and F1 scores (\textit{F1}).}}
 \label{tab:classif2}
 {
  \begin{tabular}{lccc}
   \toprule
       &\textit{Ground truth} & \textit{FaceGen} & \textit{FairGenderGen}\\
       \cmidrule[0.4pt]{1-4}
                \textit{Acc.} / \textit{F1} & 90.18\% / 90.21\% & 80.69\% / 80.76\% & 48.61\% / 47.55\%\\
   \bottomrule
  \end{tabular}
 }
\end{table} 

To eliminate gender-based distinctions in generated non-verbal behaviors, the distributions of male and female behaviors were brought closer together. As a consequence, our classifier is now less effective at distinguishing between the two, misclassifying a significant proportion of male behaviors as female. The following section examines the proximity of these distributions and quantifies the performance difference between \textit{FaceGen} and \textit{FairGenderGen}.

\vspace*{-1mm}
\subsection{Performance evaluations}
To evaluate the \textit{FairGenderGen} model, we generate videos for the two individuals, male and female, who compose the \textit{TestSet}. This involves generating all the segments and averaging overlapping image frames. Our evaluation is based on eight full videos for the objective evaluation (Section \ref{obj}) and four 30-second portions for the subjective evaluation (Section \ref{subj}). 

\vspace*{-1mm}
\subsection{Objective evaluation}\label{obj}
Objective measurements, relying on algorithmic methodologies, provide numerical performance indicators. We use mainly Dynamic Time Warping \textit{DTW}, an algorithm for measuring similarity between two temporal sequences, which may vary in speed.

\vspace*{-1mm}
\subsubsection*{\textbf{Distance between males and females}}
First, \textit{DTW} is employed to assess the similarity between the distributions of male and female non-verbal features across \textit{ground truth} data, \textit{FaceGen}-generated data, and \textit{FairGenderGen}-generated data.
For each feature, the \textit{DTW} is computed between the corresponding male and female distributions. An overall gender bias measure is obtained by averaging these \textit{DTW} distances. 

Table \ref{tab:dtwGender} confirms that the gender bias, \textit{i.e.} the distance between the two distributions, is increased using the \textit{FaceGen} model compared to the \textit{ground truth}. Generative models are capable of amplifying biases existing in the data they are trained on. However, we manage to reduce the distance between these two distributions using the \textit{FairGenderGen} model.

\begin{table}[h]
 \centering
 \setlength\tabcolsep{3pt}
 \caption{\textit{DTW} distance between males and females across \textit{Ground truth}, \textit{FaceGen} and \textit{FairGenderGen} -- {\normalfont\it The global average.}}
 \label{tab:dtwGender}
 {
  \begin{tabular}{lccc}
   \toprule
       & \textit{Ground truth} & \textit{FaceGen} & \textit{FairGenderGen} \\
       \cmidrule[0.4pt]{1-4}
                \textit{DTW} & 31.58 & 32.37 & 24.70\\
   \bottomrule
  \end{tabular}
 }
\end{table}

\vspace*{-1mm}
\subsubsection*{\textbf{Distance between the \textit{ground truth} and the generated behaviors}}
Second, \textit{DTW} is used to assess the distance between the \textit{ground truth} distributions and the generated distributions. Table \ref{tab:dtw} indicates that the distributions of \textit{FairGenderGen} are slightly further away from the \textit{ground truth} than those of \textit{FaceGen}, which may lead to a reduction in quality. We add the DTW between a static SIA (central position, AUs at intensity 0) and \textit{ground truth} for additional comparison.

\begin{table}[h]
 \centering
 \setlength\tabcolsep{3pt}
 \caption{\textit{DTW} distance between \textit{ground truth} and a static SIA, generated distributions for \textit{FaceGen} and \textit{FairGenderGen}-- {\normalfont\it The global average.}}
 \label{tab:dtw}
 {
  \begin{tabular}{lccc}
   \toprule
         & Static SIA & \textit{FaceGen} & \textit{FairGenderGen} \\
       \cmidrule[0.4pt]{1-4}
                \textit{DTW} & 29.00 & 14.18 & 14.98 \\
   \bottomrule
  \end{tabular}
 }
\end{table} 

While divergence between generated distributions and \textit{ground truth} was expected due to the intended transformation, we have to estimate whether such divergence remains within acceptable limits. Objective measures, while valuable, are insufficient since they neglect the coherence between behaviors and speech, privileging statistical similarity over contextual relevance \cite{kucherenko2023evaluating}. Consequently, to evaluate the acceptability of this divergence revealed in objective measures, subjective evaluations play a crucial role.

\vspace*{-1mm}
\subsection{Subjective evaluation}\label{subj}
To conduct subjective studies, we selected four approximately 30-second speech sequences, two featuring a female and two featuring a male speaker. These sequences were chosen semi-randomly, ensuring coherence in speech over the 30-second duration. Utilizing the \textit{Greta} platform \cite{pelachaud2015greta}, we played these sequences on SIAs, employing a male agent for non-verbal behaviors accompanying a male speech and a female agent for those accompanying a female speech. For the study setup, we employ the interface of \citet{delbosc2023towards}, inspired by other interfaces widely used in the field of behavior generation. Participants were tasked with evaluating two criteria across the four sequences: believability and temporal coordination with speech of the SIAs' behaviors. Thirty French participants, recruited on social media (15 males, 15 females, mean age 42.7, std 13.4), evaluated the two criteria through direct questions:
\begin{itemize}
    \item[o] believability: how human-like do the behaviors appear?
    \item[o] temporal coordination: how well does the agent's behavior match the speech? (In terms of rhythm and intonation) 
\end{itemize} 

Participants rated each video on a scale from 0 (worst) to 100 (best) for both criteria. The believability criterion was evaluated without sound, while the temporal coordination criterion was evaluated with sound. The results are presented in Table \ref{tab:believability}. The spectrum of responses reflects variances not just between conditions, but also includes external factors like variations in individual preferences.

Statistical analysis is carried out to examine significant differences between the \textit{FaceGen} and \textit{FairGenderGen} models, but also between male and female behaviors in these models. Initially, the normality of the data is evaluated using the Shapiro-Wilk test, confirming that the data originate from a normally distributed population. Consequently, a repeated ANOVA is used.

\begin{table}[h]
 \centering
 \setlength\tabcolsep{3pt}
 \caption{Results for the believability (Bel.) and coordination (Coo.) criteria -- {\normalfont\it Average score and standard deviation: mean (std).}}
 \label{tab:believability}
 {
  \begin{tabular}{lccc}
   \toprule
       & \multicolumn{1}{c}{All} & \multicolumn{1}{c}{Female} & \multicolumn{1}{c}{Male}\\
       \cmidrule[0.4pt]{1-4}
                \textit{Ground truth} Bel. & 53.70 (15.39) & \textbf{55.27} (15.59) & 52.13 (18.76) \\
                \textit{FaceGen} Bel. & 46.31 (15.71) & \textbf{49.88} (15.70) & 42.73 (19.36) \\
                \textit{FairGenderGen} Bel. & 49.63 (14.65) & 45.43 (15.94) & 5\textbf{3.83} (16.46) \\
                \cmidrule[0.4pt]{1-4}
                \textit{Ground truth} Coo. & 47.42 (16.05) & \textbf{51.42} (18.27) & 43.42 (15.84)  \\
                 \textit{FaceGen} Coo. & 43.33 (14.23) & 42.85 (16.84) & \textbf{43.81} (15.82)  \\
                 \textit{FairGenderGen} Coo. & 54.22 (17.69) & \textbf{54.32} (18.84) & 54.13 (18.57)  \\
   \bottomrule
  \end{tabular}
 }
\end{table}

\vspace*{-1mm}
\subsubsection*{\textbf{Perceived believability}}
There is no evidence of a decline in perceived believability of non-verbal behaviors generated by \textit{FairGenderGen} (Table \ref{tab:believability}). Statistical analysis indicates no significant difference in perceived believability with \textit{FaceGen} ($p>0.1$).

However, \textit{FairGenderGen}'s male non-verbal behaviors are significantly rated higher than \textit{FaceGen}'s male non-verbal behaviors ($p=0.008$). \textit{FairGenderGen} improves the perceived believability of male behavior. Without being significant, \textit{FairGenderGen}'s female non-verbal behaviors tends to be rated lower than \textit{FaceGen}'s female non-verbal behaviors ($p>0.1$).

In addition, by looking at the contrast in perceived believability between males and females, there is no significant difference in the perceived believability of male and female non-verbal behaviors for the \textit{ground truth} and \textit{FaceGen}. But there is significant differences for \textit{FairGenderGen}'s: where male non-verbal behaviors are rated significantly higher than their female counterparts ($p=0.029$).

\vspace*{-1mm}
\subsubsection*{\textbf{Perceived coordination}}
\textit{FairGenderGen} is significantly better than \textit{FaceGen} ($p < 0.001$). We also note that, \textit{ground truth} female's behaviors are perceived more coordinated than male's ($p=0.011$), a difference that disappeared in the generated behaviors for both \textit{FaceGen} and \textit{FairGenderGen} (Table \ref{tab:believability}). This result shows that there is no decline in performance in terms of coordination of the non-verbal behaviors generated with \textit{FairGenderGen}.

\vspace*{-1mm}
\section{Discussion and future work} \label{conclusion}
Our study highlights a new issue in the field of automatic generation of facial non-verbal behaviors: gender bias. While previous work focused mainly on the believability and coordination of these behaviors with speech, our research highlights the importance of considering the differences in non-verbal behaviors between males and females, differences already observed in real life.

We confirmed, through our analysis with a real-world dataset and the training of a state-of-the-art model in the domain, that gender biases are present in \textit{ground truth} behaviors, as well as in generated behaviors. In this paper, we have proposed a new model, \textit{FairGenderGen}, aiming to mitigate these biases and create non-verbal behaviors independent of the speaker's gender.

Our results show that \textit{FairGenderGen} effectively reduces the gender bias present in the data, even fooling a gender classifier that now recognizes much non-verbal behaviors as female's ones. The subjective evaluation shows that there is no performance loss for this model in terms of perceived coordination. However, our study also reveals a major challenge: the perception of the believability of the generated non-verbal behaviors.

Society has higher expectations of women when it comes to non-verbal behavior. For example, \citet{deutsch1987smile} revealed that the absence of a smile can be detrimental to a woman's image compared with that of a man, while there is no significant difference in image perception between smiling men and smiling women. Society's expectations of non-verbal behaviors negatively influence the perception of women who don't adopt them. 

Our efforts to mitigate gender bias in generated non-verbal behaviors resulted in a notable disparity in perceived believability performance between males and females. Female non-verbal behaviors, generally considered more believable than male ones, became significantly less believable compared to their male counterparts. We believe that these results are due to the higher stereotypical expectations placed on female non-verbal behaviors.

The disparity in perception between males and females raises essential questions for the future of research in this field. Should we direct our efforts towards maintaining stereotypes of non-verbal behaviors to preserve equivalent perceived believability and coordination between men and women, thus reflecting reality? Or should we prioritize an approach aimed at reducing these biases, even at the risk of diminishing the perception of believability?

These reflections are not limited solely to gender biases but could be extended to other sensitive variables such as cultural or racial differences. Exploring these questions more deeply could one day enable us to find answers and develop more equitable and inclusive solutions in the field of automatic generation of non-verbal behaviors.

\begin{acks}
This project was provided with computer and storage resources by GENCI at IDRIS thanks to the grant 2024-AD011014211 on the supercomputer Jean Zay's the V100 partition.
\end{acks}


\bibliographystyle{ACM-Reference-Format}
\bibliography{sample-base}

\end{document}